\documentclass[11pt,a4paper]{article}
\usepackage[utf8]{inputenc}
\usepackage[hyperref]{acl2018}
\usepackage{mathptmx}
\usepackage{latexsym}

\usepackage{url}

\usepackage{adjustbox}
\usepackage{tabularx, booktabs}

\usepackage{graphicx}
\usepackage{amssymb}
\usepackage{amsmath}
\usepackage{epstopdf}
\usepackage{subcaption}
\usepackage{xcolor}
\usepackage{metalogo}
\usepackage{tikz}
\usepackage{enumitem}

\aclfinalcopy 
\def\aclpaperid{***} 

\usetikzlibrary{arrows, decorations.markings, decorations.pathmorphing, positioning}
\usetikzlibrary{fit}					
\usetikzlibrary{backgrounds}	
\usepackage{pgfgantt}
\usetikzlibrary{mindmap,backgrounds,trees}
\usetikzlibrary{positioning,shapes,shadows,arrows}

\tikzstyle{state}=[rectangle,rounded corners,
                                    thick,
                                    minimum size=1.2cm,
                                    draw=blue!80,
                                    fill=blue!20]

\tikzstyle{input}=[rectangle,rounded corners,
                                                thick,
                                                minimum size=1.2cm,
                                                draw=orange!80,
                                                fill=orange!25]

\tikzstyle{output}=[rectangle,rounded corners,
                                    thick,
                                    minimum size=1.2cm,
                                    draw=purple!80,
                                    fill=purple!20]

\tikzstyle{matrx}=[rectangle,
                                    thick,
                                    minimum size=1cm,
                                    draw=gray!80,
                                    fill=gray!20]

\tikzstyle{noise}=[circle,
                                    thick,
                                    minimum size=1.2cm,
                                    draw=yellow!85!black,
                                    fill=yellow!40,
                                    decorate,
                                    decoration={random steps,
                                                            segment length=2pt,
                                                            amplitude=2pt}]

\tikzstyle{background}=[rectangle,
                                                fill=gray!10,
                                                inner sep=0.2cm,
                                                rounded corners=5mm]

\title{Semantic Relatedness and Taxonomic Word Embeddings}

\author{Magdalena Kacmajor\\ 
Innovation Exchange\\
IBM Ireland\\
\texttt{magdalena.kacmajor@ie.ibm.com}\\\And
John D. Kelleher\\ 
ADAPT Research Centre\\
Technological University Dublin\\
\texttt{john.d.kelleher@tudublin.ie}\\\AND
Filip Klubi\v{c}ka\\ 
ADAPT Research Centre\\
Technological University Dublin\\
\texttt{ filip.klubicka@adaptcentre.ie}\\\And
Alfredo Maldonado\\ 
Underwriters Laboratories\\
\texttt{alfredo.maldonado@ul.com}
}

\date{}

\begin{document}
\maketitle
\begin{abstract}
This paper\footnote{Recently, researchers from our lab were invited to present at an industry event on the work in the lab focusing on taxonomic word embeddings. This paper is primarily intended as a support document for this presentation. All authors contributed equally to this paper and so authorship is listed in alphabetical order.} connects a series of papers dealing with taxonomic word embeddings. It begins by noting that there are different types of semantic relatedness and that different lexical representations encode different forms of relatedness. A particularly important distinction within semantic relatedness is that of thematic versus taxonomic relatedness. Next, we present a number of experiments that analyse taxonomic embeddings that have been trained on a synthetic corpus that has been generated via a random walk over a taxonomy. These experiments demonstrate how the properties of the synthetic corpus, such as the percentage of rare words, are affected by the shape of the knowledge graph the corpus is generated from. Finally, we explore the interactions between the relative sizes of natural and synthetic corpora on the performance of embeddings when taxonomic and thematic embeddings are combined.
\end{abstract}

\section{Introduction}

Deep learning has revolutionised natural language processing over the last decade.  A key enabler of deep learning for natural language processing has been the development of \emph{word embeddings}. One reason for this is that deep learning intrinsically involves the use of neural network models and these models only work with numeric inputs. Consequently, applying deep learning to natural language processing first requires developing a numeric representation for words. Word embeddings provide a way of creating numeric representations that have proven to have a number of advantages over traditional numeric representations of language. 

Traditionally, most natural language processing used a one-hot representation for a word. Using a standard one-hot representation for a vocabulary involved first defining a vector that was the length of the vocabulary and assigning each word in the vocabulary an index in the vector. This assignment of words too indexes could be done randomly. For example the word \emph{cat} might be assigned the index 9, whereas the word \emph{dog} might be assigned the index 1. Once each word had been assigned an index then each word could be represented using a sparse vector the length of the vocabulary where all of the elements of the vector are zero apart from the index of the word the vector represents, which is set to 1 (hence the term one-hot encoding). For example, assuming a vocabulary of fifteen words, and assuming that the word \emph{cat} has an index of 9 and \emph{dog} and index of 1, then a one-hot representation would represent these words as follows:

	\begin{equation*}
       cat = [0~0~0~0~0~0~0~0~0~1~0~0~0~0~0]
       \end{equation*}
       \begin{equation*}
	dog = [0~1~0~0~0~0~0~0~0~0~0~0~0~0~0]
	\end{equation*}

Although conceptually simple, these one-hot representations have a number of drawbacks. First the dimensionality of the representation is the size of the vocabulary, and given that we may often want to develop models including tens or hundreds of thousands of words, these representations can become very unwieldy. Not only that, but they are also very sparse, so much of the memory allocated to storing these large representations is wasted. Second, such a representation does not naturally encode any semantic relationships between words: for example all words are equally similar and dissimilar i.e. the cosine distance between all word vectors is the same. 

A word embedding also uses a vector of numbers to represent a word. However, word embeddings are dense distributed representations (as distinct from the sparse localist representation of the one-hot representation) \cite{kelleher:2019}. Consequently, word embedding representations generally have a much lower-dimensionality in comparison with one-hot: typically using only several hundred or low thousands of dimensions. This lower dimensionality makes them much more computationally tractable from a memory perspective. Furthermore, a number of different approaches---see \emph{inter alia} \cite{bengio_neural_2003,collobert_scratch_2011,mikolov2013,pennington2014glove}---have been developed to learn the embedding vectors for words directly from a natural language corpus. Typically, using these algorithms the vector assigned to a word positions (or embed) the word in a multi-dimensional space so that the relative sposition of the word in the embedding space with respect to other words encodes semantic relationships between that word and other words. For example, the vectors assigned to the names of European capitals may position these terms in a similar region in the embedding space (thereby encoding their semantic similarity) or male and female versions of terms may be located at the same offset from each other. Indeed, \citet{mikolov2013} demonstrated this relative encoding of semantics using the example that subtracting the vector for \emph{man} from the vector for \emph{king} and then adding the vector for \emph{woman} to the result results in a vector similar to vector for \emph{queen}:
\begin{equation*}
vec(King) - vec(Man'') + vec(Woman') \approx vec(Queen)
\end{equation*}

The standard linguistic philosophy underpinning much of the work on learning embeddings is the distributional hypothesis that is sometimes summarised as \emph{you shall know the meaning of a word by the company it keeps} \cite{firth1968synopsis}. This approach defines semantic similarity in terms of: two words are semantically similar if they appear in similar contexts. Consequently, the processes for learning embeddings can learn that \emph{London} and \emph{Paris} are semantically similar because both of these words occur in contexts that are often applied to European capital cities; for example, either word might occur in the following contexts: ``the government in \_\_\_\_ announced'', ``\_\_\_\_ is one of the largest cities in Europe'', ``the European summit in \_\_\_\_'', ``\_\_\_\_ is only two hours flight from Madrid''.  However, this form of semantic relatedness is not the only form of semantic relatedness that language expresses, or relies upon. This observation opens up a number of questions, such as: What types of semantic relatedness does language encode? and, Can we train embeddings to encode other types of semantic relatedness?

This paper is structured into three main sections with each section providing a partial synopsis of a primary publication. Section \ref{sec:thematicVtaxonomic} primarily draws on \cite{Kacmajor2019}; Section \ref{sec:taxonomicembeddings} is based on \cite{klubicka2019}; and Section \ref{sec:datasetsize} is based on \cite{maldonado2019size}. The goal of this paper is to provide an overarching narrative across these publications that highlights the linkages across the reported experiments. Section \ref{sec:thematicVtaxonomic} introduces distinctions between a number of different types of semantic relatedness, in particular distinguishing between thematic and taxonomic relatedness, and highlights that the standard forms of different families of models of lexical semantics capture different forms of semantic relatedness. Building on the distinction between thematic and taxonomic relatedness, Section \ref{sec:taxonomicembeddings} focuses on taxonomic embeddings: these are embeddings that have been trained so that the relative position of words in the embedding space encodes relatedness within a taxonomy. One well known method for training taxonomic embeddings is to use random walks over a taxonomy to generate a pseudo-corpus and then to training embeddings using this psuedo-corpus.  The experiments reported in Section \ref{sec:taxonomicembeddings} analyse how the shape of the taxonomy affects the properties of the generated pseudo-corpus and the consequent embeddings trained on the corpus. Section \ref{sec:datasetsize} focuses on embeddings that integrate taxonomic and thematic information and how the relative sizes of the respective corpora can affect overall quality of the final embeddings. Finally, in Section \ref{sec:conclusions} we review and link the work presented in the paper and discuss potential future directions for research. 

\section{Semantic Relatedness: Thematic versus Taxonomic Relatedness}
\label{sec:thematicVtaxonomic}

One of the contributions by \citet{Kacmajor2019} is to define, and clarify the distinctions between, a number of terms that are often found in the literature on semantic relatedness, specifically they propose the following set of definitions: 

\begin{description}
	\item[\textbf{Semantic relatedness}] the broadest category that comprises any type of semantic relationship between two concepts.
	\item[\textbf{Taxonomic relations}] a subset of relatedness defined as belonging to the same taxonomic category, which involves having common features and functions. In the literature, this type of relatedness is often referred to as \textbf{similarity}.
	\item[\textbf{Non-taxonomic relations}] relatedness existing by virtue of co-occurrence of concepts in any sort of context.
	\item[\textbf{Thematic relations}] a subset of non-taxonomic relations defined as co-occurrence in events or scenarios, which involves performing \textit{complementary roles}.
\end{description}

The primary  distinction in semantic relatedness is between \emph{taxonomic} and \emph{non-taxonomic} relatedness with \emph{thematic} relatedness being particular form of non-taxonomic relatedness.  The distinction between these different forms of relatedness can be understood from a number of different perspective, most notably at the conceptual (cognitive) level and at the linguistic level. Focusing on the conceptual level concepts that are taxonomic related share features, \emph{river} and \emph{brook} are taxonomically related because they are both bodies or water; whereas, concepts that a non-taxonomically related may not shared features but are instead considered related because they often co-occur, consider the example of a \emph{spider} and a \emph{web}. From a linguistic perspective the distinction between taxonomic and non-taxonomic relatedness can be understood as analogous to the distinction between \emph{paradigmatic selection} and \emph{syntagmatic combination} within language. Finally, within the class of non-taxonomic relatedness the distinction characteristic of thematic relatedness is that not only do the concepts co-occur but that they also play complementary roles within the situations and context they occur in. For example, the concepts of \emph{doctor} and \emph{nurse} are thematically related because they play complementary roles within surgery. Table \ref{tab:conceptualAndlinguistic} lists these different levels of distinctions between the different forms of relatedness. 

\begin{table}
\begin{center}
\begin{tabular}{c|c|c}
~ & Taxonomic & Non-Taxonomic\\
\hline
Conceptual & Shared Features & Co-occurrence\\
Linguistic &Paradigmatic Selection & Syntagmatic Combination\\
Thematic & ~ & Complementary Roles$^\star$\\
\hline
\end{tabular}
\end{center}
\caption{Conceptual and linguistic distinctions between taxonomic and non-taxonomic relatedness}
\label{tab:conceptualAndlinguistic}
\end{table}

Both forms of semantic relatedness are important in NLP. For example, in language modelling the accuracy of a language models is dependent on the ability of the model to successfully learn the co-occurrence dependencies between words, even if these words co-occur at long-distances \cite{salton2017attentive,mahalunkar2018using,mahalunkar2018understanding,salton2019}. So from this perspective a language model must be able to learn the \textbf{thematic} relationships between words because these relationships express high-probability of co-occurrences and thus help predict the next word. However, for NLP applications such as paraphrasing it is useful to know when one word can be replaced by another, and so in this application \textbf{taxonomic relations} become important. 

\subsection{Models of Semantic Relatedness}

Models of semantic relatedness can be broadly categorised as being knowledge based or distributional. Using a knowledge based model the semantic relatedness between to terms is calculated based on the distance between the terms in a hand-crafted knowledge graph; wordnet based distance metrics are prototypical examples of this type of model. By contrast, distributional models are calculate semantic relatedness in terms of co-occurrence statistics. In vector distributed models, such as word2vec \citet{mikolov2013}, these co-occurrence statistics get encoded in terms of vector similarity and as such semantic relatedness is often measured in terms of the cosine distance between vectors. 

The most direct method for evaluating a model of semantic relatedness is to compare the semantic relatedness estimates generated by a model for a given pair of words against gold standard estimates provided by human judgments. This comparison between model estimates and human judgements is typically done in terms of a Spearman rank correlation with respect to one or more datasets. There are a number of datasets that are suitable for use in such an evaluation. Some of these datasets mainly capture thematic relatedness (e.g., the evocation dataset \citep{Boyd-Graber:Fellbaum:Osherson:Schapire-2006}  the thematic relatedness norms dataset \citep{jouravlev_thematic_2015}) whereas other datasets capture thematic relatedness (e.g.,  the Simlex-999 dataset \citep{Hill2014} and the USF Free Association Norms (\url{http://w3.usf.edu/FreeAssociation})). \citet{Kacmajor2019} report experiments comparing the performance of knowledge based and distributional models across these datasets and found that the relative performance of models of different types across these datasets varied systematically, specifically: knowledge-based models do badly on the thematic dataset, whereas distributional models do badly on the taxonomic datasets. There are a number of findings that can be taken from these results, and the other results reported in \citet{Kacmajor2019}; however, for the purposes of this paper the main finding we highlight is that vanilla distributional models are limited in their ability to capture taxonomic relations. Given the current popularity of distributional models, particularly the distributed versions of these models, it is worthwhile to consider if there are ways in which distributional models can be adapted to capture taxonomic relatedness. 

\section{Learning Embeddings from Wordnet}
\label{sec:taxonomicembeddings}

\cite{goikoetxea2015} proposed an approach to created distributed (vector based) representations of term semantics that capture taxonomic semantic relatedness based on the idea of using a random walk over a WordNet to generate a synthetic corpus and then using standard distributional based methods, such as word2vec, to train word embedding vectors. This is a somewhat elegant solution to the challenge of how to train taxonomic word embeddings, the use of a random walk to generate the synthetic corpus results in the co-occurrence statistics within the synthetic corpus reflecting taxonomic relatedness (as encoded in the knowledge graph the random walk traverses) rather than thematic relatedness. Consequently, standard models, such a word2vec, can then be used to learn taxonomic word embeddings without any alteration.

Although elegant we identified some gaps in the literature in relation to the analysis of this approach; in particular, in relation to the analysis of the properties of the synthetic corpora generated using the random walk. \cite{klubicka2019} reports on experiments to analyse the properties of these synthetic corpus. A particular focus of this research was the relationship between the shape of the underlying knowledge graph and the properties of the synthetic corpora. The results of the experiments indicate that when the knowledge graph has a tree-shape structure then the synthetic corpora exhibit similar scaling properties to those of natural corpora, such as Zipf's law and Heaps' law. One reason why this is interesting is that it indicates that the synthetic corpora generated by random walks over tree-shaped knowledge graphs will likely include rare words. This is relevant to semantic similarity work because rare words in the synthetic corpora may affect the performance of embeddings on word similarity evaluations. And, in fact \cite{klubicka2019} also report experiments that demonstrated the impact of rare words on the performance of word embeddings. The insight that rare words are present in synthetic corpora is useful because it can provides some information in relation to answering the question: how large should a pseudo-corpus be? 

\section{Dataset Size for Taxonomically Enriched Word Embeddings}
\label{sec:datasetsize}

\cite{maldonado2019size} directly address the question of how large a psuedo-corpus should be, as part of a wider set of experiments that examined the performance of representations that combine both taxonomic and thematic information. As an example to provide some background to these experiments, although the random-walk based method proposed by \cite{goikoetxea2015} generates word embeddings that capture taxonomic relatedness we would ideally like to have representations that capture both thematic and taxonomic relatedness. Previous work suggests that performance improvements on both forms of relatedness can be achieved by combining models of different types. For example, \cite{Kacmajor2019} report results for an array of representations that were created by concatenating different models together. In general there are two different methods for combining thematic and taxonomic information: (i) create separate taxonomic and thematic representations and then concatenate them together, or (ii) train a representation on a thematic corpus and then fine-tune it to a particular domain using a synthetic taxonomic corpus. Within this context, \cite{maldonado2019size} explored how the relative sizes of the thematic and taxonomic corpora used to train the models affect the final overall performance of the resulting representations. The paper reports on a large number of experiments testing different combinations of models and datasets. There are a number of finding reported in the paper, however the primary results with respect to taxonomic embeddings are that: (a) taxonomic enrichment of thematic embeddings (i.e., embeddings that are originally trained on natural corpora) through vector combination (either via concatenation or fine-tuning) does not always improve the performance of the embeddings and generally works best when the natural corpus is large; and (b) where performance does increase, only medium sizes of random walk corpora are required - i.e., in the context of models that combine thematic and taxonomic information, there is little benefit to training vectors on very large random walks. 

\section{Conclusions}
\label{sec:conclusions}

One of the key advantages of word embeddings for natural language processing is that they encode the semantic relatedness between words. However, as \cite{Kacmajor2019} highlights there are different types of semantic relatedness and different lexical representations can encode different forms of relatedness. We have analysed one approach to learning word embeddings that encode taxonomic relatedness based on training the embedding on a synthetic corpus that has been generated via a random walk over a taxonomy. Through a range of experiments we demonstrate how the properties of the synthetic corpus, such as the percentage of rare words, are affected by the shape of the knowledge graph the corpus is generated from \cite{klubicka2019}. And, so when using a random walk techniques understanding the shape of the graph, and also the connectivity of the graph, are useful in order to estimate the size of the synthetic corpus required to address the problem of rare-words. Furthermore, it is possible to combine taxonomic embeddings with embeddings from natural corpora and for this combination to work well; however, when doing this care should be taken with respect to the size of the respective corpora \cite{maldonado2019size}. This article is intended to provide an overview of a theme of work in our lab that provides a connection between the above papers. In doing this we have highlighted the aspects of each of the above papers that relate to this connecting narrative. However, each of the above papers provides a broader set of experiments and results than the ones highlighted in this paper, and also a deeper and more contextualised analysis of the results. 

In future work we will focus more on sentence level embeddings, which have been receiving a lot of interest in recent years. For example, machine translation typically works at a sentence level, and often struggles with the processing of idiomatic phrases \cite{salton2014empirical}. Interestingly, \cite{salton2016idiom,salton2017idiom} demonstrated that sentence embeddings encode information relating to the usage of idiomatic phrases in a sentence thereby showing that sentence level embeddings have the potential to help machine translation systems address the problem of idomaticity in translation. Using a very similar methodology to \cite{salton2016idiom}, \citet{conneau2018you} have proposed the concept of a probing task to understand what types of information sentence level embeddings encode. Our immediate future work will focus on understanding sentence level embeddings through the design and use of a variety of probing task experiments.

\section*{Acknowledgements}

This research was supported by the ADAPT Research Centre. The ADAPT Centre for Digital Content Technology is funded under the SFI Research Centres Programme (Grant 13/RC/2106) and is co-funded under the European Regional Development Funds.

\bibliography{AA-Taxonomic-Embeddings}
\bibliographystyle{acl_natbib}

\end{document}